# GraftNet: An Engineering Implementation of CNN for Fine-grained Multi-label Task

Chunhua Jia, Lei Zhang, Hui Huang, Weiwei Cai, Hao Hu, Rohan Adivarekar

*Abstract*—Multi-label networks with branches are proved to perform well in both accuracy and speed, but lacks flexibility in providing dynamic extension onto new labels due to the low efficiency of re-work on annotating and training. For multi-label classification task, to cover new labels we need to annotate not only newly collected images, but also the previous whole dataset to check presence of these new labels. Also training on whole re-annotated dataset costs much time. In order to recognize new labels more effectively and accurately, we propose GraftNet, which is a customizable tree-like network with its trunk pretrained with a dynamic graph for generic feature extraction, and branches separately trained on sub-datasets with single label to improve accuracy. GraftNet could reduce cost, increase flexibility, and incrementally handle new labels. Experimental results show that it has good performance on our human attributes recognition task, which is fine-grained multi-label classification.

## I. INTRODUCTION

MULTI-LABEL classification is a fundamental task in computer vision area which has achieved great progress due to the development of deep convolutional networks. Basically there are three kinds of solutions.

The first one transfers multi-label task into a set of binary classifications for each single label[23,1]. But there is redundancy because of repeatedly extracting some general features which can probably be extracted once with one model. Although this solution is optimized by techniques such as regularization constrains and label encoding[25, 26, 3, 4], it is still not straightforward. The second solution uses a transfer-learned network to extract features and independent classifiers such as SVM are then applied to predict for each label. It is more often explored and utilized in multi-label classification in past years. CNN is a much more powerful feature extractor, by combining off-the-shelf CNN with SVM the computation cost can be reduced,but problems still exist, as CNN-SVM is a two-step process and is less accurate compared to full CNN models. The third way is partly the same as the second one, it uses an end-to-end convolutional network[11, 14] to treat multi-label prediction as one complete task, which is much more straightforward. In this kind of end-to-end solution, tree-like network is a typical architecture that uses shallow layers to extract low-level common features and then uses separate branches to extract high-level features for each label[24]. Because of its good classification performance and high computational efficiency, tree-like networks are much used in recent days.

However, tree-like networks mentioned above are designed for a static dataset with fixed number of labels and can not effectively deal with the situation where the dataset labels increase dynamically and quick response for new labels is required. They will consume too much resource in annotation and training for the newly increased labels. Specifically two problems emerge when a new label is incoming. First is to annotate not only the new images collected for this new label but also all the existing images to identify whether they belong to this new label or not. As we cannot know the labels to be predicted in advance, the whole dataset must be re-annotated again and again for newly increased labels. When the number of labels keeps increasing and the dataset increases to hundreds thousands or millions level, this could cost much labor and time. Besides, training on the whole dataset all over again is also costly. Above two problems prevent us from quickly recognizing new labels with an end-to-end network.

In this paper, we abandon to maintain a whole dataset where each image is annotated with all the labels, and explore training multi-label networks only on sub-datasets with single label.We fuse the second and the third way to propose GraftNet, a new architecture for multi-label classification task, which fixes issues mentioned above and fits applications in practice better. GraftNet is also a tree-like network but customizable. Its trunk is for generic feature extraction and is pretrained with a dynamic data flow graph. Each branch is trained separately on a sub-dataset corresponding to one label. By training trunk and branches in a two-step way, GraftNet could actually save time and labor of annotation and training especially for dynamically increasing labels' scenario. Besides, training task of one-branch-for-one-label (including samples collection and model fine-tuning -- the whole loop) is more practical and manageable in practice. Experimental results show that GraftNet performs well on our human attributes recognition task, i.e. fine-grained multi-label classification, and the combination of pretrained trunk and fine-tuned branches can effectively improve the accuracy.

## II. RELATED WORKS

Target of multi-label classification is to predict multiple labels for each input image. In past, multi-label classification was usually achieved by transforming multi-label problem into a set of single label problems. The basic technique is Binary Relevance, which trains one binary classification model for each label[23, 1]. Then hard parameter sharing method[24] appeared on stage that is similar to tree model. More concretely in [11], AlexNet which is pretrained on the ImageNet was used to extract features and SVM was used to train for each label, and in [14], the network was fine-tuned by target multi-label dataset to learn more domain-specific features.

However, this way is limited and ineffective when number of labels needed to predict is large, and ignores relations and co-occurrence among labels. For example, a male couldn't be pregnant and a maintenance worker has higher possibility to be a male than female. For the purpose of refining predicted results, many efforts in multi-label classification area were put into capturing label relations. Ranking-based learning strategy that transforms multi-label problem into label ranking problem and label space encoding, i.e. classifier chain, that adds the predicted result of last label to input embedding of next model, were designed to consider correlations among labels so that performance of multi-label classification can be improved when intra-class information exists[2,3,4,5,6]. Furthermore, attention models designed for sequence prediction such as recurrent neural networks (RNNs), long short-term memory (LSTM) and so on, were used to encode labels into embedding features and thus making the relationship among labels better exploitable[7,8]. Later, graph methods were introduced to obtain semantic relations. Hierarchy and Exclusion(HEX) graph was involved in exploring correlations of labels, which captures semantic relations by using mutual exclusion, overlap and subsumption [10]. To improve the performance, spatial regularization was introduced to capture both semantic and spatial relations[8,9], knowledge graphs[12] and Graph Convolutional Network (GCN)[13] were utilized to learn inter-dependent information.

Nevertheless, multi-label classification is not only used to describe whole image attribute, but also used to describe local attribute. For example, each person's attribute could be blurred by global representation for pictures containing many people, and person's detailed attribute such as face, hat, body shape, bags and so on, cannot be represented accurately by global features of the pictures. In order to extract local features, objects proposals[15, 16] that includes possible foregrounds were applied.

Even if those ways achieve relatively better performance by exploiting co-occurrence and correlations among labels and extracting better representation both in global and in local, it still takes long time and lots of resources to train a complete deep neural network for each label. Specifically in this paper, we put forward GraftNet, which is more resource-saving and flexible, especially for incremental extension on new labels.

## III. GRAFTNET

### 3.1 Preprocess: Instance segmentation of each individual passenger

Considering occlusion is inevitable when there are several passengers in the elevator at the same time, instance segmentation is applied to segment each passenger. Different from object detection, instance segmentation can accurately segment all objects at pixel level and will minimize the impact of occlusion and background. It could be considered as a preprocess of attention mechanism, so that our model for human attributes recognition could focus on human target itself completely.

Similar to object detection that principally includes two-stage methods[18, 15] and one-stage methods[16, 17], instance segmentation also has two-stage models[19, 20] focusing on higher accuracy and one-stage models to speed up segmentation. Mask RCNN[19] is the most popular method in two-stage instance segmentation that combines Faster RCNN and Fully Convolutional Network(FCN) to segment object at pixel level. However, though accuracy of localization is increased, pooling features by ROI Align is a serial computing process that makes Mask RCNN hard to accelerate. As a result, YOLACT[21], a representative one-stage method, was proposed to speed up instance segmentation. YOLACT adds an FCN branch on top of the backbone to produce k prototype masks and additionally predict k mask coefficients for each anchor. The prototype masks and the mask coefficients are then multiplied together to produce masks for each object. YOLACT is faster due to one-stage design and can generate masks with high quality.

We choose YOLACT to segment person and remove background and the occuluded parts of other non-target person in images as shown in Fig 1. The original open source implementation is based on PyTorch, and we have re-implemented it with Tensorflow in order to deploy using Tensorflow Serving. The segmented person crops are collected to constitute the sub-dataset with single label for each attribute. In our paper, we collect 16 sub-datasets for 18 attributes, each sub-dataset contains positive and negative samples except that two datasets contain 3 classes such as gender dataset labeled with 3 classes including Male, Female, and Other, and all of the sub-datasets are independent with each other. We then try to train GraftNet only on these sub-datasets.

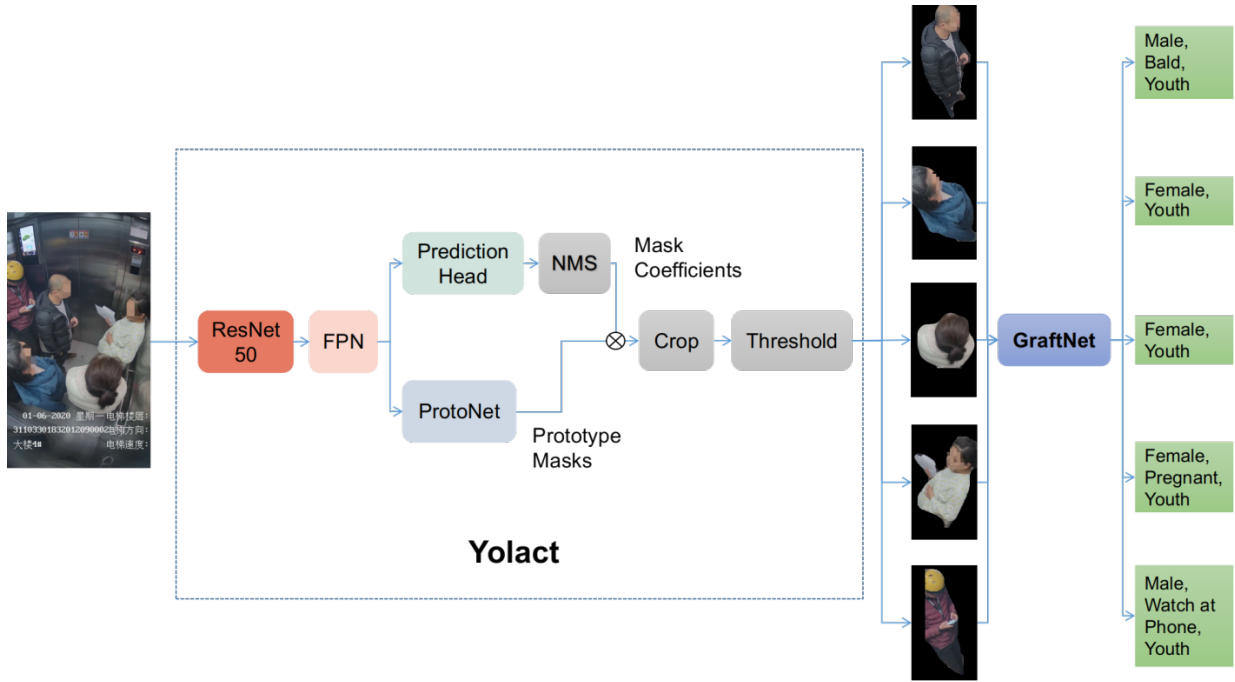

Fig. 1. Architecture of YOLACT [21].

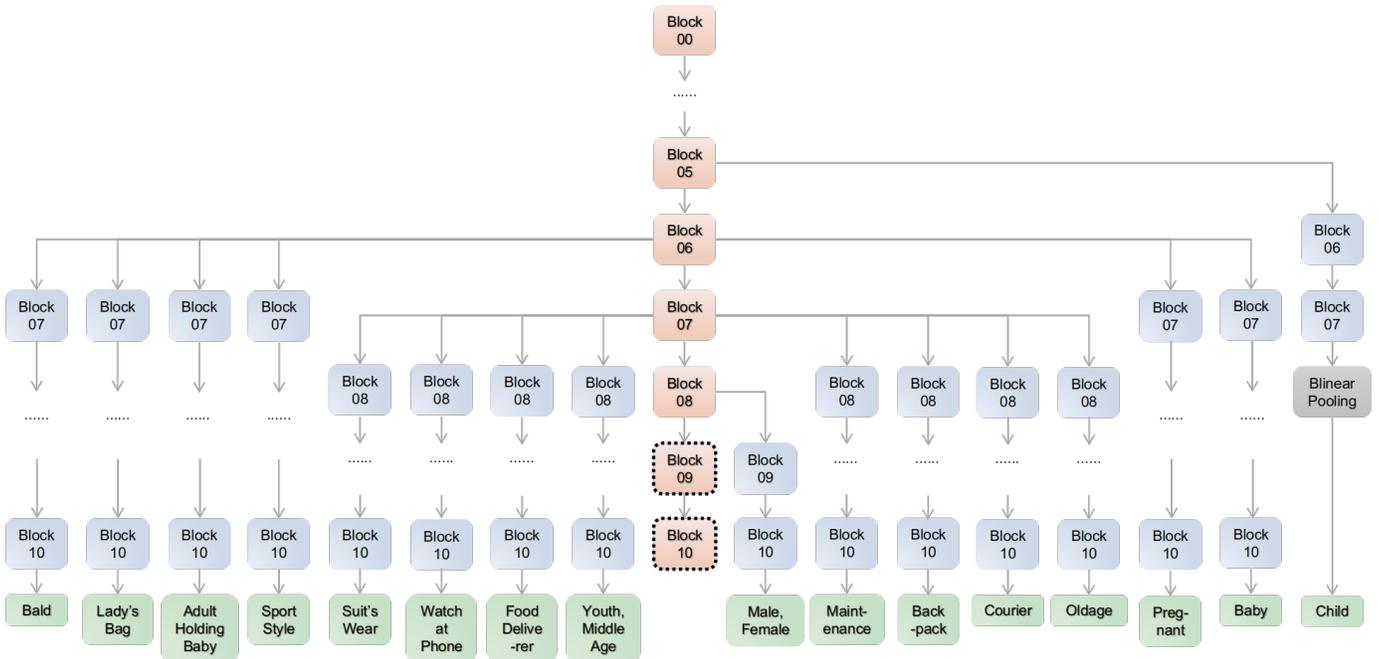

Fig. 2. Architecture of GraftNet.

### 3.2 Architecture

To recognize human attributes, low-level features such as shapes and textures could be commonly represented with generic features and high-level features should be customized for different attributes, therefore GraftNet, a tree-like network, is designed as shown in Fig 2. To construct and train GraftNet on the sub-datasets, we propose a two-step training procedure. In first step, InceptionV3 [32] is pretrained on the collection of the sub-datasets by using a dynamic data flow graph, which will be described in Sec 3.2 (InceptionV3 is consisted of 11 blocks, as for now all our grafting operations are block basis). The blocks with pretrained weights (pink ones) could be considered as the trunk of GraftNet. The second step is to separately graft and fine-tune branches (blue ones) on the trunk for each attribute.

As shown in Fig 2, trunk (pink blocks) is for low-level generic feature extraction and are shared with all branches during inference, and branches (blue blocks) extract high-level features specifically for each attribute. To simplify dataset management and reduce cost, we collect and annotate sub-datasets for single attribute only, branches are trained on sub-datasets correspondingly and then grafted onto the trunk. Regarding the structure of branches, for most of them we simply follow the original structure of InceptionV3, meaning the fine-tuning is based on the InceptionV3 with the pretrained weights of the trunk, we only have to decide how many consecutive blocks to be set as trainable according to the actual result. More blocks can be opened for fine-tuning if the attribute is difficult to recognize. A little exploration we did is to reduce blocks and apply some extension on some branch, which will be introduced in Sec 3.4. More explorations as such could be done in next.

Since GraftNet is deployed on cloud but not embedded, we focused more on extendability rather than efficiency. From some perspective, our work is quite like a reverse process compared to network pruning or compression. Rather than to reduce the redundancy of neural networks for a fixed task, what we did is to leverage the over-parameterization and maximum its usage with a few extra branches. We have to admit that due to engineering purpose the implementation of GraftNet is a little rough; for example, the grafting operation is block basis and the final optimized architecture (i.e. grafting branches) is based on experimental results but not methodology. In the future, more sophisticated techniques such as networks design, weights compression, etc., can be involved and methodology can be developed for networks extendability and over-parameterization leverage.

### 3.3 The Trunk: Pretrained weights by using dynamic graph

Transfer learning is a common way to train model for a new task and helps to accelerate training convergence and achieve better accuracy, especially when the target dataset is relatively small. Usually the pretrained weights for transfer-learning are gathered from the training on some public dataset, e.g. ImageNet [33], COCO[34], etc., which is tremendously large and thus the weights are generic for many tasks. Recently,research[22] pointed out that transfer-learning benefits more from pretraining on a source domain that is more similar to target domain. In our case, target datasets, i.e. sub-datasets, are bunches of images of one person with one attribute labeled. First thing is that clearly there is low domain similarity between our target datasets and public dataset such as ImageNet. And second, to combine all the sub-datasets as whole one could make the domain similarity high enough because they are actually the same, but all the images have to be annotated for all the attributes once again. Furthermore, for dynamically increasing labels' scenario, we have to re-work on the whole dataset over again when a new label is incoming. In practice it's unaffordable for us.

To overcome above challenges, we design a graph with dynamic data flow in pretraining step to utilize the sub-datasets directly without any rework. This training process aims to make the pretrained weights as much generic to all the attributes as possible, but not to directly achieve better accuracy comparable to each independent branch. As shown in Fig 3, our dataset is composed of single sub-dataset containing negative and positive images samples of each attribute. In each training step, one batch of images is randomly selected from one sub-dataset and fed into InceptionV3 together with their labels and attribute ID of this batch. Global Average Pooling(GAP) [35] is applied to flatten features and followed by one set of parallel placed FC layers corresponding to each attribute to get outputs. Condition op activates only one FC layer corresponding to the attribute ID input within each batch. Only the weights of Conv. and Batch Normalization (BN) layers of InceptionV3 will be taken as trunk for fine-tuning of each branch and those FC layers will be excluded. We were not sure if training with such dynamic graph would converge at the very beginning, yet experiment showed so.

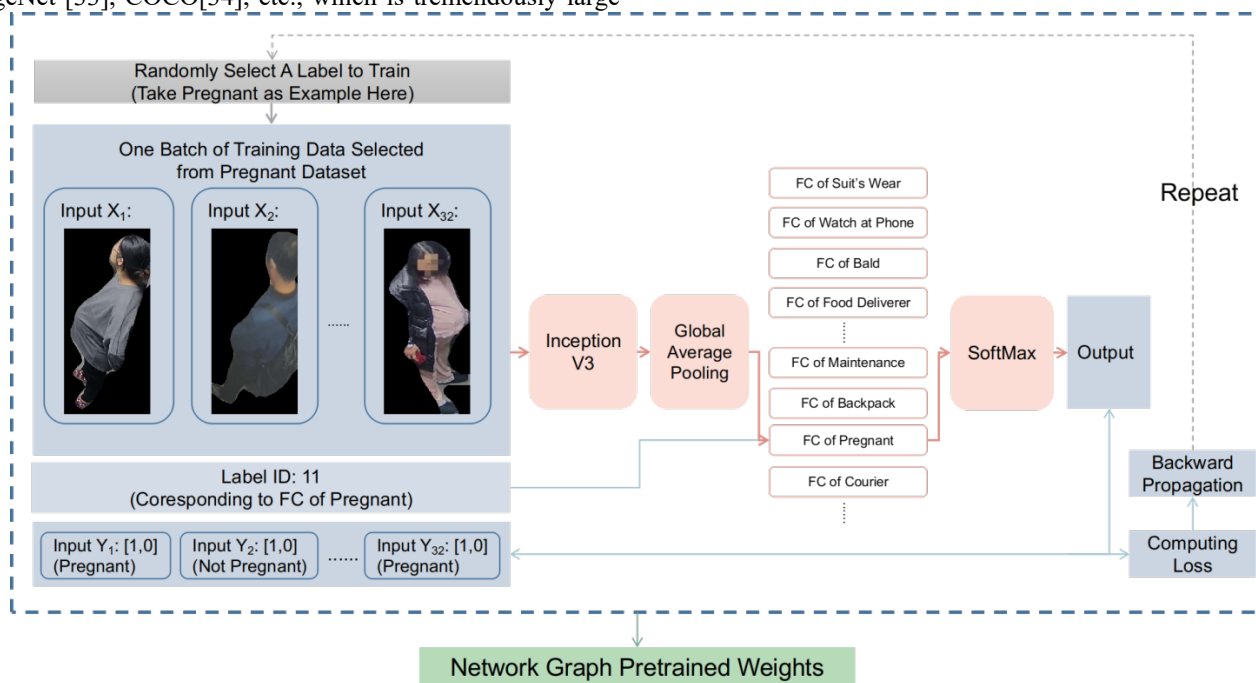

Fig. 3.Process of Training with Dynamic Data Flow Graph.

## 3.4 Hard negative samples mining

Additionally, in practice some negative samples are hard to be distinguished from positive, thus we continuously collect such negative samples, which makes the number of negative much larger than positive and the sub-dataset imbalanced. Inspired by [22], to optimize the distribution of positive and negative samples, we apply Earth Mover's Distance (EMD)[30, 31] to keep the hard negative samples and remove easy ones. Specifically, we use our trunk model to extract a 2048d vector for each image in the dataset of an attribute, cluster the negative samples into several clusters by k-means, and calculate the EMD and similarity between each cluster and the positive samples. The clusters that are close to the positive are kept as hard negative samples and part of other clusters are removed to reduce redundancy.

## 3.5. Training frequency of trunk and branches

As the trunk is trained jointly on the collection of the sub-datasets by dynamic data flow graph, it usually costs too much time for every training. To reduce cost, the trunk is trained once after 5-7 new labels are accumulated. At the beginning we only had 5 labels, when the number of labels increased to 10, we re-train the trunk on the 10 sub-datasets and update the branches. The same is true for 15 labels. As for branches, once a new label is incoming a new branch is trained on the corresponding sub-dataset and added to the trunk. Of course the trunk is robuster if train more labels.

## 3.6 Adjustment and extension on branch for further improvement

Based on the architecture of trunk and branches, the branches could be pruned and some tricky techniques of fine-grained image classification[27, 28, 29] can be utilized on some branches depending on the actual requirement and effect. This makes GraftNet more flexible and customizable. For example, as shown in Fig 2, to recognize "child" better and faster, we take only first 8 blocks of InceptionV3 (blocks 0~5 are from the trunk and blocks 6~7 are open for fine-tuning as part of the branch) to extract features and use Bilinear Pooling[28] to improve accuracy. Experiment shows network with less blocks and Bilinear Pooling has better generalization ability. Furthermore, any appendable methods and techniques with better effect can be effortlessly put into use because of the flexibility of this architecture.

## 3.7 Application: real-time advertising system using computer vision

We build a prototype of real-time advertising system which is partly based on our implementation of elevator passenger attributes recognition. LED screen and camera are both mounted in each elevator with Internet access. YOLACT and GraftNet are deployed on cloud to capture attributes of elevator passenger from uploaded camera snapshot and then advertisements are pushed onto LED screen accordingly. For now, there is no actual recommendation algorithm for ads pushing. The pushing is based on the schema we make from business logic and experience. For example, for someone holding a little baby, we push ads of diaper or toy. Recommendation algorithm would be explored in our work next.

## IV. EXPERIMENT

### 4.1 Dataset

We only collect independent datasets for each attribute. The trunk is trained on these datasets jointly by the proposed dynamic graph, and then the branches are fine-tuned separately on each dataset. All the images are collected via the cameras installed in 70 thousand elevators. We use these cameras to get the video screenshots of the elevators and upload and save the screenshots to a web server, we then download the saved screenshots according to demands. The resolution of the screenshots is 1920*1080 or 1280*720, these screenshots are then processed by YOLACT to segment the samples for training and testing.

As we only need to train on some independent datasets, each of these datasets only has tens of thousands of images and it is much easier to collect and annotate. Number of dataset images for each attribute are shown in the table below. Besides, we don't collect any identification information that may invade privacy of individuals.

TABLE 1
COMPARISON OF SCORES OF USING DIFFERENT PRETRAINED WEIGHTS

| Attributes | Number of train images: pos/neg | Number of test images: pos/neg |
|---|---|---|
| Suit's Wear | 20876/96092 | 2486/9998 |
| Watch at Phone | 116531/199291 | 1226/19582 |
| Bald | 57511/116619 | 1105/5957 |
| Food Deliverer Person | 56647/143042 | 5382/10658 |
| Lady's Bag | 14713/80263 | 672/19328 |
| Youth | 84330/50833 | 10523/16774 |
| Middle Age | 85815/49348 | 6552/20745 |
| Adult Holding Baby | 27907/187138 | 2287/79205 |
| Sport Style | 12983/91529 | 5433/57303 |
| Maintenance | 26181/86701 | 10323/22173 |
| Backpack | 28697/81504 | 7400/40000 |
| Pregnant | 8131/106533 | 903/11837 |
| Courier | 26181/86701 | 4768/18173 |
| Oldage | 99347/525590 | 23889/127822 |
| Baby | 46348/166021 | 2707/79205 |
| Male | 180275/177706 | 47636/48344 |
| Female | 192147/165834 | 37909/58071 |
| Child | 110147/510938 | 31770/119927 |

### 4.2 Training

The training process includes training the trunk and fine-tuning the branches. We first use dynamic graph to train an InceptionV3 to get a base-InceptionV3 network, this base-InceptionV3 will act as trunk for all the branches. When new attribute demand comes and the new sub-dataset is col-lected, we start to train a new InceptionV3 on the small sub-dataset. For this step of training process, we first load the pretrained weights of the previous trained base-InceptionV3, freeze specific number of blocks and only train the remaining unfrozen blocks. We experiment several choices to determine which blocks to freeze or to train. After training, we detach the unfrozen layers from this newly trained InceptionV3 and attach them to base-InceptionV3 as a separate branch. When other new demand comes, the same steps will be repeated. The trunk will be trained once or twice a month and the training process usually takes 4-5 days. The new InceptionV3 trained on small sub-dataset will be trained when new demand comes and it usually takes 1-2 days. This training manner helps us in quickly responding to new demands.

*4.3 Results*

Experimental results are shown in the figures below, ROC curves for each attribute are computed. We first analyze the overall performance of GraftNet with graph network pretrained weights, and then test GraftNet with ImageNet pretrained weights to verify the proposed graph network with dynamic data flow procedure.

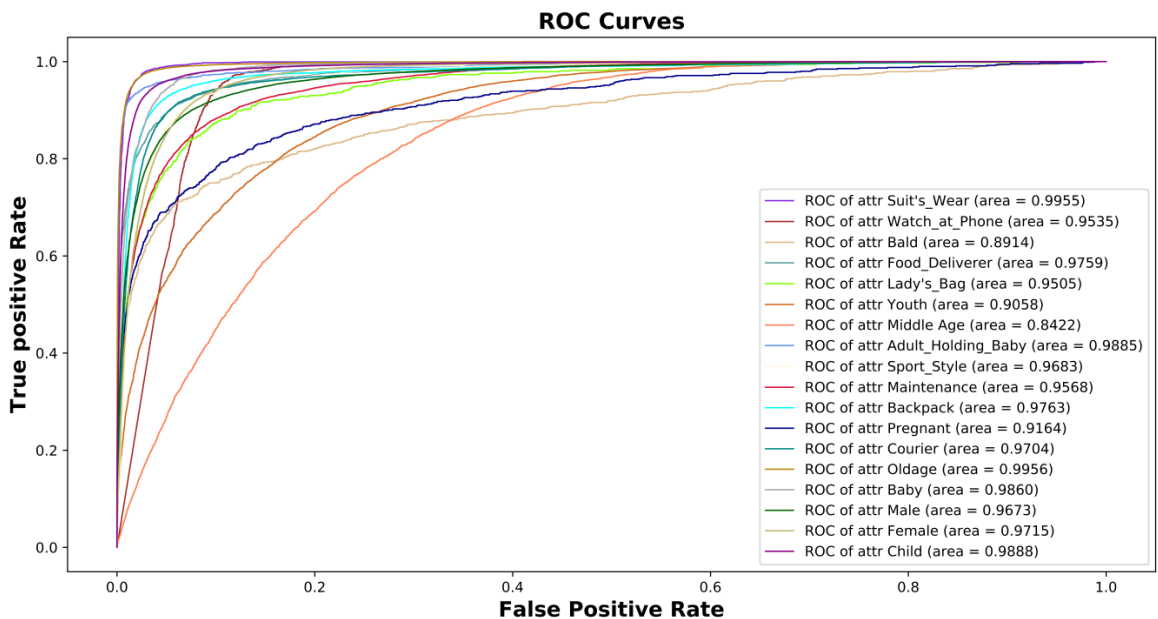

Fig. 4.ROC curves of attributes with our pretrained weights.

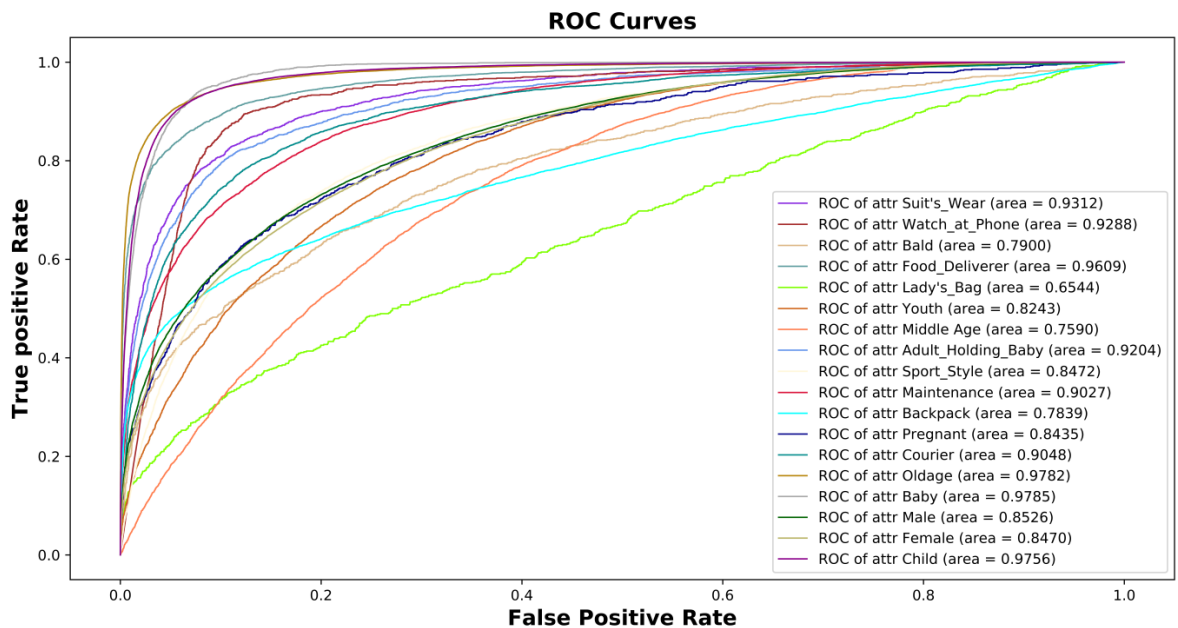

Fig. 5.ROC curves of attributes with pretrained weights on ImageNet.

### 4.3.1 Overall classification results of GraftNet

Results of GraftNet are shown in Fig 4 and Fig 5, there are totally 18 attributes in our experiment. We compute the ROC curves and corresponding AUC scores for each attribute. As can be seen, GraftNet performs well on most of the attributes and their AUC scores are between 0.95-0.99. This means the construction process of GraftNet and the network graph with dynamic data flow are effective for multi-label classification and can achieve excellent performance. But there are still four attributes of bald, youth, middle-age and pregnant which do not perform well enough, we think this is because the features of these attributes are hard to discriminate. For attributes of bald and pregnant, the appearance features are not obvious, and for attributes of youth and middle-age, it's hard to draw a clear line between them.

### 4.3.2 Classification results by using different kinds of pre-trained weights

To verify the effectiveness of the proposed network graph with dynamic data flow, we trained GraftNet with ImageNet pretrained weights additionally and the test results are shown in Fig 5. As can be seen from Fig 4 and Fig 5, ROC curves of Fig4 performs much better than Fig 5 and most of the AUC scores of Fig 5 are among 0.7-0.9. The attribute of lady's bag performs the worst and the AUC score is only 0.65, but with graph network its AUC score improves to 0.95. So our graph network with dynamic data flow is effective to improve the classification accuracy.

### 4.3.3 Best scores for each attribute

By selecting threshold with the best scores, 18 attributes' evaluation metrics including threshold, accuracy, recall, and False-Positive Rate, are shown in Table 1. It can be seen that all the classification models using weights pretrained on our own dataset with dynamic graph perform better than models using weights pretrained on ImageNet.

TABLE 2
COMPARISON OF SCORES OF USING DIFFERENT PRETRAINED WEIGHTS

|    | Results of Weights Pretrained on Our Own Dataset | | | | | Results of Weights Pretrained on ImageNet | | | | |
|----|--------|--------|--------|--------|--------|--------|--------|--------|--------|--------|
|    | THR | Acc. | Prec. | Rec. | FPR | THR | Acc. | Prec. | Rec. | FPR |
| 00 | 0.9400 | 0.9777 | 0.9339 | 0.9553 | 0.0168 | 0.9900 | 0.8993 | 0.7936 | 0.6681 | 0.0432 |
| 01 | 0.9999 | 0.9793 | 0.8920 | 0.8210 | 0.0080 | 0.9900 | 0.9736 | 0.9081 | 0.7188 | 0.0058 |
| 02 | 0.0010 | 0.9159 | 0.8198 | 0.5928 | 0.0242 | 0.9500 | 0.8709 | 0.6852 | 0.3231 | 0.0275 |
| 03 | 0.9990 | 0.9223 | 0.9484 | 0.8127 | 0.0223 | 0.9000 | 0.9074 | 0.9200 | 0.7930 | 0.0348 |
| 04 | 0.9990 | 0.9739 | 0.6729 | 0.4315 | 0.0073 | 0.9300 | 0.9671 | 0.6327 | 0.0461 | 0.0009 |
| 05 | 0.0100 | 0.8225 | 0.7520 | 0.8053 | 0.1668 | 0.1000 | 0.7484 | 0.6806 | 0.6546 | 0.1928 |
| 06 | 0.9000 | 0.7916 | 0.5808 | 0.4744 | 0.1081 | 0.9900 | 0.7653 | 0.5463 | 0.1321 | 0.0347 |
| 07 | 0.5000 | 0.9914 | 0.8465 | 0.8465 | 0.0044 | 0.9900 | 0.9707 | 0.4797 | 0.5155 | 0.1610 |
| 08 | 0.9990 | 0.9425 | 0.8055 | 0.4428 | 0.0101 | 0.9900 | 0.9151 | 0.5438 | 0.1222 | 0.0097 |
| 09 | 0.9990 | 0.8844 | 0.9155 | 0.7008 | 0.0301 | 0.9000 | 0.8397 | 0.7948 | 0.6677 | 0.0802 |
| 10 | 0.9400 | 0.9564 | 0.8691 | 0.8485 | 0.0237 | 0.9000 | 0.8872 | 0.8048 | 0.3666 | 0.0165 |
| 11 | 0.4000 | 0.9555 | 0.8206 | 0.4762 | 0.0079 | 0.4000 | 0.9377 | 0.7280 | 0.1927 | 0.0055 |
| 12 | 0.9999 | 0.9392 | 0.8752 | 0.8253 | 0.0309 | 0.9999 | 0.8684 | 0.8364 | 0.4558 | 0.0234 |
| 13 | 0.9000 | 0.9800 | 0.9598 | 0.8937 | 0.0070 | 0.2000 | 0.9570 | 0.8841 | 0.8194 | 0.0200 |
| 14 | 0.8500 | 0.9801 | 0.6966 | 0.7041 | 0.0105 | 0.9900 | 0.9747 | 0.6180 | 0.6154 | 0.0130 |
| 15 | 0.5000 | 0.9122 | 0.9117 | 0.9113 | 0.0870 | 0.9000 | 0.7609 | 0.8115 | 0.6749 | 0.1544 |
| 16 | 0.8000 | 0.9198 | 0.8886 | 0.9111 | 0.0745 | 0.2000 | 0.7707 | 0.7284 | 0.6687 | 0.1628 |
| 17 | 0.4000 | 0.9760 | 0.9201 | 0.9042 | 0.0208 | 0.4000 | 0.9550 | 0.8790 | 0.8207 | 0.0300 |

*00-Suit's Wear, 01-Watch at Phone, 02-Bald, 03-Food Deliverer Person, 04-Lady's Bag, 05-Youth, 06-Middle Age, 07-Adult Holding Baby, 08-Sport Style, 09-Maintenance, 10-Backpack, 11-Pregnant, 12-Courier, 13-Oldage, 14-Baby, 15-Male, 16-Female, 17-Child.

### 4.3.4 Result of the extension on branch

To test the effect of Bilinear Pooling on the branch of child, we further train and test the branch of child with normal setting, which is using 11 blocks and traditional average pooling. Experimental result is shown in Table 3, we select the predicted top ~3300 positive images according to classification scores and static the false positive images in them. For Bilinear Pooling there are only 410 false positive images, while without Bilinear Pooling there are 850 false positive images. As a result, Bilinear Pooling is helpful to improve attribute classification accuracy.

TABLE 3
RESULTS OF BILINEAR POOLING AND GLOBAL AVERAGE POOLING

|  | Top positive images | False positive images |
|---|---|---|
| InceptionV3 block 10 + Global Average Pooling | 3344 | 850 |
| InceptionV3 block 07 + Bilinear Pooling | 3307 | 410 |

### V. CONCLUSION

In this paper, we have proposed GraftNet with its trunk extracting low-level features and branches extracting high-level features that is suitable for fine-grained multi-label classification task in industry implementation. We formulate dynamic graph to pretrain trunk of GraftNet. During the training procedure a complete dataset labeled with all attributes is not required and therefore more efforts are saved from annotation. And flexible branches attached to trunk extract customized high-level features to corresponding attributes. Compared to previous method, it is able to reduce time spent on annotation and training, simplify multi-label classification, enhance speed of reaction to predicting new label, and still maintain relatively high accuracy of fine-grained tasks. Experimental results prove the ascendancy of our framework.